%% file: main.tex
\documentclass[10pt,twocolumn,letterpaper]{article}

\usepackage{cvpr}              
\input{preamble}

\definecolor{cvprblue}{rgb}{0.21,0.49,0.74}
\usepackage[pagebackref,breaklinks,colorlinks,allcolors=cvprblue]{hyperref}

\def\paperID{29076} 
\def\confName{CVPR}
\def\confYear{2027}

\title{HAVANA: A Hierarchical Video-Action Model for Vision-Language Navigation}

\author{
Dongzhihan Wang$^{1,2}$ \quad
Yi Du$^{2}$ \quad
Chen Wang$^{2}$ \quad
Xiuxian Li$^{1,\dagger}$ \quad
Liang Xu$^{3, \dagger}$ 
\\[4pt]
$^{1}$Tongji University \qquad
$^{2}$SAIR Lab \qquad
$^{3}$Shanghai University
\\[2pt]
\\[-5pt]
{\small $^{\dagger}$Corresponding authors}
}

\begin{document}
\maketitle

\begin{strip}
    \centering
    \vspace{-15mm}
    \includegraphics[width=\linewidth]{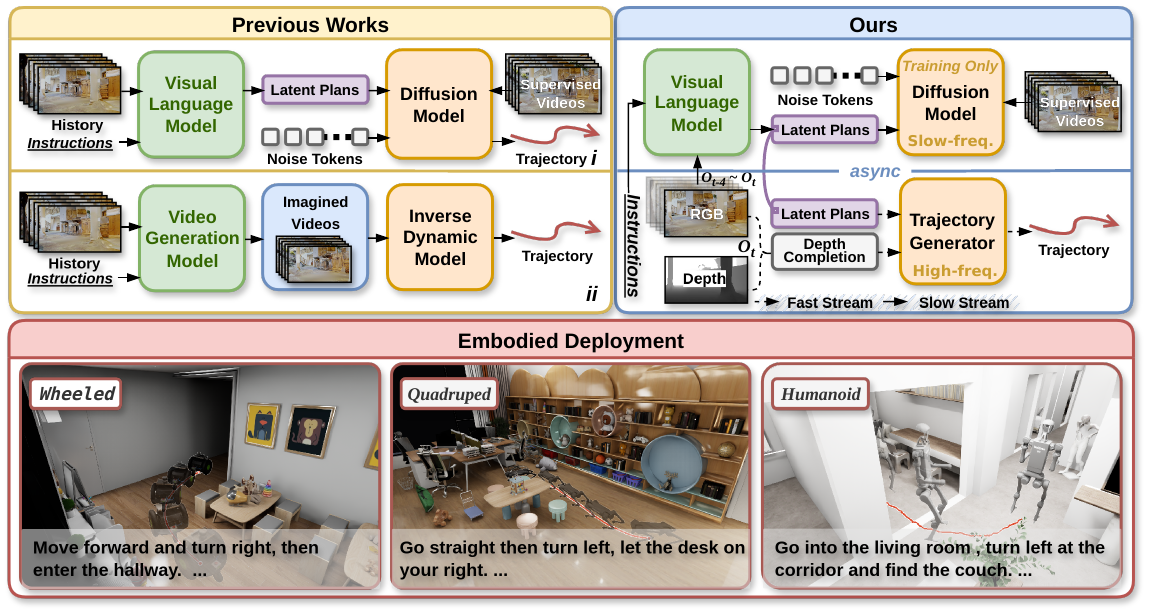}
    \captionsetup{hypcap=false}

    \captionof{figure}{\textbf{HAVANA separates predictive reasoning from reactive local trajectory prediction.} The illustrated \textcolor{teaseryellow}{previous works} include two designs: joint generation of future videos and trajectories (\romannumeral 1), and trajectory recovery from generated videos via inverse dynamics (\romannumeral 2). \textcolor{undblue}{HAVANA} learns compact instruction-conditioned latent plans through history observations. Its two systems run asynchronously, allowing each plan to guide multiple local trajectory updates from fresh RGB-D observations. Future-video generation is omitted during inference. \textcolor{wayred}{Experiments} demonstrate zero-shot navigation generalization across diverse embodiments and scenarios.}
    \captionsetup{hypcap=true}
    \label{fig:intro}
\end{strip}

\input{sec/0_abstract}    
\input{sec/1_intro}

\input{sec/2_related_work}
\input{sec/3_method}
\input{sec/4_experiments}
\input{sec/5_conclusion}
{
    \small
    \bibliographystyle{ieeenat_fullname}
    \bibliography{main}
}


\end{document}

%% file: preamble.tex
\usepackage{cuted}
\usepackage{caption}
\usepackage{multirow}
\usepackage{siunitx}
\usepackage{tabularx}
\usepackage{booktabs}
\usepackage{multirow}
\usepackage{array}
\usepackage{algorithm}
\usepackage{algpseudocode}
\usepackage{booktabs, multirow, xcolor, colortbl}

\definecolor{undblue}{HTML}{2D5F94}
\definecolor{genorange}{HTML}{9A5800}
\definecolor{wayred}{HTML}{A12F31}
\definecolor{teaseryellow}{HTML}{D6B656}

\usepackage{etoolbox}

\makeatletter
\patchcmd{\abstract}
  {\vspace*{12pt}}
  {\vspace*{2pt}}
  {}{}
\makeatother

%% file: sec/0_abstract.tex
\begin{abstract}
Vision-Language Navigation (VLN) requires long-horizon instruction-consistent planning and short-horizon path planning grounded in the currently observed geometry. Anticipating how future observations evolve along the instructed route provides predictive guidance beyond the current field of view, motivating video-action models (VAMs) that learn representations of future visual states. However, many existing VAMs couple future visual prediction with action generation during inference, despite their distinct temporal requirements: instruction-conditioned route guidance can be reused across multiple control steps, whereas local trajectories must continuously adapt to fresh observations. To bridge this gap, we introduce \textbf{HAVANA}, a dual-system VAM that decouples predictive reasoning from reactive local planning.  A slow system produces a compact language-conditioned latent plan from observation history, while a fast system combines the latest plan with the current observation to generate local trajectories. Their asynchronous execution enables rapid adaptation without repeated future prediction, and no future video is generated during inference. Experiments on OmniNavBench demonstrate state-of-the-art navigation performance over prior baselines, while zero-shot real-world deployments further show robust generalization to unseen environments.
\end{abstract}

%% file: sec/1_intro.tex
\section{Introduction}
\label{sec:intro}
Vision-Language Navigation (VLN) requires an embodied agent to follow natural-language instructions through sequential interaction with its environment. Recent approaches~\cite{Zhang2025UniNaVidAV,cheng2025navila,Wei2025StreamVLNSV} directly predict actions or local navigation goals from instructions and observation histories. This leaves future route consistency implicit: when the next instruction-relevant landmark is not yet visible, a locally plausible action does not guarantee reaching the scene expected by the instruction. Learning to predict future semantics provides an additional supervision for this correspondence and has been shown to improve navigation on longer routes \cite{li2023improving}.

Video-based World-Action Models (WAMs), also referred to as Video-Action Models (VAMs)~\cite{pai2025mimicvideo,Yuan2026FastWAMDW,wang2026unified}, introduce predictive capability by learning future visual states together with actions. For VLN, however, long-horizon instruction-consistent planning and short-horizon local path planning require different information at different rates. To follow instructions beyond the current field of view, long-horizon planning requires semantic reasoning about future observations and route progression, whereas local trajectory generation needs to continuously adapt to latest observed geometry to avoid obstacles. This conflict leaves open how those representations should be organized for navigation across different temporal scales.

Existing fast-and-slow VLN frameworks~\cite{wei2026ground, zhu2026fsd} demonstrate the benefits of separating reasoning from execution. They typically condition a trajectory policy on latent goals and current observations, with its planning interface learned through pixel-goal grounding and trajectory supervision. We instead investigate how future visual prediction can ground a reusable planning interface in anticipated route progression. Such instruction-conditioned guidance can remain useful across multiple trajectory updates, while each local trajectory prediction must account for newly observed geometry. We therefore ask: \textbf{how can low-frequency future visual prediction guide high-frequency trajectory generation from fresh observations?}

We introduce HAVANA (\textbf{H}ier\textbf{A}rchical \textbf{V}ideo-\textbf{A}ction Model for -----\textbf{NA}vigation), a dual-system framework that reuses predictive latent plans for local trajectory prediction. The slow system uses a pretrained Vision-Language Model (VLM) to encode observation history and language instructions into planning states. Joint future-video and long-horizon waypoint supervision grounds these states in visual evolution and intended motion. The lightweight fast system combines the latest latent plan with the current RGB-D observation to generate short-horizon trajectories, which a tracking controller converts into velocity commands. Both systems run asynchronously, allowing the fast policy to incorporate new observations between planning updates while reusing latent route guidance, without video generation.

Our contributions can be summarized as follows:

\begin{itemize}
    \item We propose HAVANA, a dual-system VAM for VLN that decouples low-frequency predictive reasoning from high-frequency reactive trajectory prediction.
    \footnote{Source code: \url{https://github.com/Anastasiawd/HAVANA}.}
    \item We learn plan representations through joint future-video and waypoint supervision, augmented with an instruction-ranking objective that encourages instruction-consistent predictions over conflicting alternatives.

    \item We transfer trajectory prediction priors learned from language-free expert data to language-conditioned navigation through predictive latent plans, enabling local trajectory prediction without video generation.
\end{itemize}

%% file: sec/2_related_work.tex
\section{Related Work}
\label{sec:related_work}

\begin{figure*}[t]
    \centering
    \includegraphics[width=\linewidth]{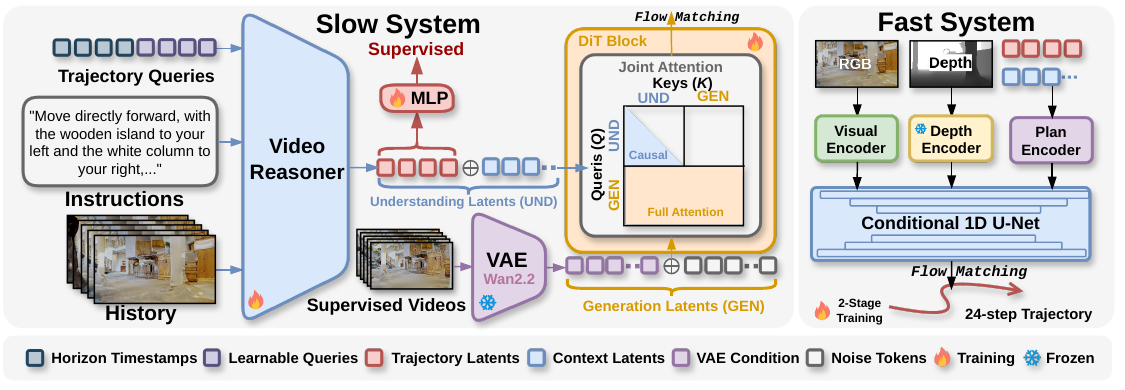}
\caption{\textbf{Overview of the HAVANA training pipeline.}
\textbf{Left:} The slow system processes an RGB observation history, a navigation instruction, and four learnable time-conditioned queries with a video reasoner. Its contextual and query states form the \textcolor{undblue}{\textbf{understanding (UND)}} stream. The query states are also decoded by a supervised MLP into four egocentric waypoints (\textcolor{wayred}{\textbf{waypoint prediction}}). A clean conditioning latent and noisy future latents form the \textcolor{genorange}{\textbf{generation (GEN)}} stream. UND uses causal self-attention; GEN jointly attends to both streams within each sample. Future latents are supervised by flow matching.
\textbf{Right:} RGB, depth, and UND latents condition a 1D U-Net for flow-matching-based prediction of a 24-step trajectory.
}
    \vspace{-1em}
    \label{fig:pipeline}
\end{figure*}

\subsection{Video-Action Models for VLN}

\noindent\textbf{Explicit Video-to-Action Models.}
Video-based planning uses generated visual futures as intermediate plans for action execution. UniPi~\cite{du2023learning} formulates control as text-conditioned video generation followed by inverse dynamics. ImagiNav~\cite{chen2026imaginav} extends this paradigm to navigation through subgoal-conditioned egocentric videos, while NavDreamer~\cite{huang2026navdreamer} selects imagined visual plans and reconstructs executable 3D waypoints. These approaches provide explicit visual foresight, but action recovery depends on the geometric consistency of generated videos. Iterative video generation also increases the cost of replanning from new observations, complicating high-frequency closed-loop control.

\noindent\textbf{Latent World-Action Models.}
World-action models (WAMs) integrate predictive visual representations with action learning. DreamZero~\cite{Ye2026WorldAM} and AstraNav-World~\cite{Hu2025AstraNavWorldWM} jointly model future visuals and actions, while WAM-Nav~\cite{yang2026wamnav} uses asymmetric joint diffusion for long-horizon actions and short-horizon visual foresight. WorldVLN~\cite{Zhao2026WorldVLNAW} decodes predicted latent world transitions into waypoints, and SparseVideoNav~\cite{Zhang2026SparseVG} reduces inference cost through sparse future prediction. Other approaches retain predictive supervision while avoiding explicit future-video generation at inference: Fast-WAM~\cite{Yuan2026FastWAMDW} uses video co-training to improve direct action prediction, whereas ForeWAM~\cite{Huang2026ForesightWS} exposes latent future context to an action decoder. AHA-WAM~\cite{Cai2026AHAWAMAsynchronousHW} separates low-frequency world planning from high-frequency action execution for robotic manipulation. These advances establish predictive representations as useful control signals. For VLN, a key design question is how instruction-conditioned visual and motion foresight can provide reusable route guidance across successive reactive control updates.

\subsection{Hierarchical VLN Frameworks}

Hierarchical navigation frameworks decouple high-level instruction-consistent reasoning from low-level path planning. At the reactive level, ViNT~\cite{Shah2023ViNTAF} learns transferable visual navigation representations, NoMaD~\cite{Sridhar2023NoMaDGM} employs diffusion policies for local trajectory generation, and FlowNav~\cite{Gode2024FlowNavCF} uses flow matching \cite{Lipman2022FlowMF}. DualVLN~\cite{wei2026ground} combines a slow VLM planner producing pixel goals and latent planning features with a fast diffusion policy \cite{Cai2025NavDPLS} for asynchronous navigation. FSD-VLN~\cite{zhu2026fsd} extends this separation to aerial navigation, while NavForesee~\cite{liu2026navforesee} combines hierarchical language planning with short- and long-horizon visual prediction. In VLN, route planning must preserve instruction intent over a longer horizon, whereas local control must adapt to current observation. HAVANA connects these roles through language-conditioned predictive latent plans learned with future video and waypoint supervision. A separately trained reactive policy reuses these plans alongside real-time RGB-D observations, enabling asynchronous control without video generation.

%% file: sec/3_method.tex
\section{Method}
\label{sec:method}
As shown in Figure \ref{fig:pipeline}, \textbf{HAVANA} is a dual-system framework that separates long-horizon semantic reasoning from short-horizon trajectory generation. The two systems operate asynchronously: the slow system continuously updates the high-level plan, while the fast system reacts to current observations using the most recently available plan. The predicted trajectory is subsequently tracked by a model-predictive path-integral (MPPI) controller \cite{Williams2017ModelPP}. This design preserves the semantic and predictive world modeling while retaining the responsiveness, geometric awareness, and closed-loop safety required for embodied navigation.

\subsection{Latent Planning with the Slow System}
Our slow system is designed around a simple observation: a navigation representation should not only describe where the agent should move, but also anticipate how the visual scene will evolve along that motion. Direct waypoint supervision provides a compact geometric learning signal, yet it does not explicitly encourage the multimodal representation to model the future consequences of a navigation decision. We therefore couple trajectory prediction with latent future-video modeling, using the latter as an auxiliary supervisory signal for learning predictive navigation representations.

Given an observation history \(\mathbf O_t\) and the global navigation instruction \(\ell\), we use a 2B multimodal reasoner \cite{Basant2025NVIDIANN} \(\mathcal R_{\psi}\) as the video reasoner. We augment its input with four learnable trajectory queries $\mathbf Q_{\mathrm{traj}}=\{q_1,q_2,q_3,q_4\}$. Before entering the reasoner, each query is augmented by a learned projection of a sinusoidal encoding of its assigned horizon at $\tau_k\in\{0.8,1.6,2.4,3.2\}\ \mathrm{s}.$  We then obtain understanding latents:
\begin{equation}
\begin{aligned}
(\mathbf C_t,\mathbf Z_t)
=
\mathcal R_{\psi}
(\mathbf O_t,\ell,\mathbf Q_{\mathrm{traj}}), 
\end{aligned}
\end{equation}
where \(\mathbf C_t\) denotes the semantic \textbf{context latents} and \(\mathbf Z_t=\{z_{t,k}\}_{k=1}^{4}\) contains \textbf{trajectory latents.} Each trajectory token is decoded by a multilayer perceptron (MLP) into an egocentric waypoint, yielding four waypoints in total. The trajectory queries are explicitly supervised by future robot poses, encouraging them to form a compact structured representation of the intended motion.

For future-video supervision, we follow the Cosmos3 \cite{agarwal2026cosmos} video-training recipe for latent future-video supervision and extend its understanding stream with our waypoint representation. All valid reasoner states, including $\mathbf Z_t$ and the context states $\mathbf C_t$, are passed to a Diffusion Transformer (DiT) \cite{Peebles2022ScalableDM} generator based on Mixture-of-Transformers (MoT) \cite{Liang2024MixtureofTransformersAS} architecture as understanding (UND) tokens. Thus, the trajectory latents are shared by waypoint prediction and video generation. In the video branch, a frozen Wan2.2 VAE \cite{wan2025} encodes a supervision clip containing the current frame and 32 future frames into nine temporal latents. The first latent $\mathbf x_0$ remains clean as the current-frame condition; the other eight $\mathbf x^{\mathrm{fut}}=\{\mathbf x_j\}_{j=1}^{8}$ are perturbed with noise and represent the future-video targets. These latents form the generation (GEN) stream. Within the DiT blocks, UND queries attend causally to UND tokens, while GEN queries attend to both UND and GEN tokens. We use conditional flow matching (CFM) \cite{ Tong2023ImprovingAG} to supervise only the noised future latents, providing an additional learning signal to the trajectory-query states used for waypoint prediction. For a sampled noise level
$\sigma$ and $\boldsymbol\epsilon\sim\mathcal N(\mathbf 0,\mathbf I)$,
we construct
\begin{equation}
\begin{aligned}
\widetilde{\mathbf x}_{\sigma}^{\mathrm{fut}}
&=(1-\sigma)\mathbf x^{\mathrm{fut}}
  +\sigma\boldsymbol\epsilon,\\
\mathbf v^{\star}
&=\boldsymbol\epsilon-\mathbf x^{\mathrm{fut}}.
\end{aligned}
\end{equation}
The generator predicts the velocity
$\hat{\mathbf v}_{\theta}$ from the clean current latent, noised
future latents, noise level, and reasoner outputs. Its slow-system CFM loss is shown below:
\begin{equation}
\begin{aligned}
\mathcal L_{\mathrm{CFM}}^{\mathrm{slow}}
=\frac{1}{B}\sum_{i=1}^{B}
 \frac{1}{|\Omega_i|}
 \sum_{u\in\Omega_i}
 \bigl(\hat v_{\theta,i,u}-v_{i,u}^{\star}\bigr)^2,
\end{aligned}
\end{equation}
where $B$ is the minibatch size, and $\Omega_i$ denotes the scalar
elements of the future-video latents for sample $i$.

To additionally encourage the visual prediction to depend on the navigation instruction, we introduce an \textbf{instruction-ranking objective.} For each sample, we evaluate the same noised video under the matched instruction and a mismatched instruction. The two passes share the observation history, noise level, noise realization, and velocity target. Let $d_i^{+}$ and $d_i^{-}$ be their respective future-latent errors, computed using the inner average in the equation above. We apply a ranking loss with a margin $\delta$:
\begin{equation}
\begin{aligned}
\mathcal L_{\mathrm{rank}}
=\frac{1}{B}\sum_{i=1}^{B}
 \left[\delta+d_i^{+}-d_i^{-}\right]_+.
\end{aligned}
\end{equation}

Finally, the complete slow-system objective is a weighted combination:
\begin{equation}
\begin{aligned}
\mathcal L_{\mathrm{slow}}
=\lambda_{\mathrm{traj}}\mathcal L_{\mathrm{traj}}
+\lambda_{\mathrm{CFM}}^{\mathrm{slow}}\mathcal L_{\mathrm{CFM}}^{\mathrm{slow}}
+\lambda_{\mathrm{rank}}\mathcal L_{\mathrm{rank}}.
\end{aligned}
\end{equation}

These objectives turn a pretrained VLM from a purely semantic planner into a future-aware trajectory reasoner: waypoint supervision grounds its latent queries in geometry, while latent video prediction teaches them to capture the visual consequences of navigation decisions.

\subsection{Trajectory Generation with the Fast System}
Our fast system (FastNav) operates as the reactive tier of HAVANA, grounding the high-level latent plan into a short-horizon trajectory under the current camera stream. FastNav processes the latest aligned RGB-D frame conditioned on the most recent slow-system plan, responding to immediate obstacles and local geometry at high frequency while preserving high-level semantic intent. 

\noindent\textbf{Plan-Conditioned Observation Encoding.} FastNav conditions each local update on the slow system's context and trajectory latents, $(\mathbf C_t,\mathbf Z_t)$. A lightweight plan encoder compresses $\mathbf C_t$ into eight context tokens and projects the four trajectory latents into ordered motion tokens. Learned embeddings of the plan age and each waypoint's remaining horizon are added to form the time-aligned plan tokens $\mathbf P_t$. EfficientNet \cite{Tan2019EfficientNetRM} encodes the current RGB frame, while a frozen LingBot-Depth \cite{Tan2026MaskedDM} backbone extracts geometric features from the frame and its aligned metric depth map. The resulting observation tokens attend to $\mathbf P_t$, yielding plan-conditioned features $\mathbf H_t$ using cross-attention \cite{Vaswani2017AttentionIA}. 

\noindent\textbf{Conditional Trajectory Generation.} FastNav predicts $K=24$ egocentric waypoints, each specifying forward displacement, lateral displacement, and heading relative to the current robot pose. We convert the target waypoints into normalized successive increments $\mathbf a$ and train a temporal 1D U-Net \cite{Ronneberger2015UNetCN} by CFM. For Gaussian noise $\boldsymbol\epsilon$ and $s\sim\mathcal U(0,1)$, the training objective is
\begin{equation}
\begin{aligned}
\mathbf a_s &= (1-s)\boldsymbol\epsilon+s\mathbf a,\\
\mathcal L_{\mathrm{CFM}}^{\mathrm{fast}}
&= \mathbb E_{s,\boldsymbol\epsilon}
\left\|v_\theta(\mathbf a_s,s\mid\mathbf H_t,\mathbf P_t)
-(\mathbf a-\boldsymbol\epsilon)\right\|_2^2 .
\end{aligned}
\end{equation}
The pooled observation features provide global conditioning, while the U-Net attends to observation and plan tokens across temporal resolutions. At inference, the predicted field is integrated from Gaussian noise; the resulting increments are accumulated to recover the local trajectory.

\subsection{Asynchronous Closed-Loop Navigation}

HAVANA decouples slow planning from fast control through a background worker with one active request and a replaceable pending request. Upon completion, the worker publishes a consistent plan snapshot. At each control step, FastNav uses the newest plan completed before the current observation was captured, without waiting for the next slow update. The robot holds until the first plan arrives. This latest-only schedule prevents outdated requests from accumulating.

We align a cached plan with the current observation using its age,
$a_t=t-t_{\mathrm{src}}$, where $t_{\mathrm{src}}$ is the timestamp of the observation that produced the plan. A time adapter adds learned embeddings of $a_t$ and the signed remaining horizons $\tau_k-a_t$ to the plan tokens. Replanning from the rolling observation history refreshes the semantic context and near-term trajectory latents as the agent moves. Plan versions and measured latency support runtime control checks but are not learned model inputs.

Given the current RGB-D observation and the latest long-horizon plan, FastNav samples local trajectory candidates from independent Gaussian initializations.
A multi-reference MPPI controller evaluates rollouts against these candidates. It scores trajectory tracking, clearance, and control regularity while rejecting swept-footprint collisions. Weighted control averages are rechecked within each reference before selection. The controller executes only the first command of a feasible rollout before replanning.

The complete navigation loop is summarized in Alg.~\ref{alg:havana_navigation}. Here $\mathcal H$ is a rolling observation buffer; $\mathrm{History}_5(\mathcal H,\mathbf o_t)$ appends the current observation and keeps only the most recent five observations, which are used as context for replanning. $\widehat{\mathcal Y}_t$ denotes the local trajectory predicted by FastNav, and $\mathbf u_t$ is the velocity command executed at time $t$.

\begin{algorithm}[t]
\caption{HAVANA Closed-Loop Navigation}
\label{alg:havana_navigation}
\begin{algorithmic}[1]
\Require Instruction $\ell$, slow planner $\mathcal S_\psi$,
         FastNav $\mathcal F_\theta$
\State $\mathcal H\gets\varnothing,\quad
       \mathcal M\gets\varnothing$
\While{episode active}
    \State $(\mathbf o_t,t)\gets\mathrm{Observe}()$
    \State $\mathcal H\gets
           \mathrm{History}_5(\mathcal H,\mathbf o_t)$
    \State $\mathcal M\gets
           \mathrm{UpdateMap}(\mathcal M,\mathbf o_t)$
    \State $(\mathbf P_t,t_{\mathrm{src}})\gets
           \mathrm{GetLatestPlan}
           (\mathcal S_\psi,\mathcal H,\ell,t)$
    \State $\mathbf u_t\gets\mathbf 0$
    \If{$\mathbf P_t\neq\varnothing$}
        \State $a_t\gets t-t_{\mathrm{src}}$
        \State $\widehat{\mathcal Y}_t\gets
               \mathcal F_\theta(\mathbf o_t,\mathbf P_t,a_t)$
        \State $\mathbf u_t\gets
               \mathrm{MPPI}(\widehat{\mathcal Y}_t,\mathcal M)$
    \EndIf
    \State $\mathrm{Execute}(\mathbf u_t)$
\EndWhile
\end{algorithmic}
\end{algorithm}

%% file: sec/4_experiments.tex
\section{Experiments}
\label{sec:experiments}
\subsection{Experimental Setup}
\input{tables/omninavbench}
\noindent\textbf{Data \& Implementation Details.} The slow system is fine-tuned on InternData-N1 \cite{interndata-n1} to provide compact latent plan representations. FastNav is trained in two stages: Stage~1 pretrains a no-goal local trajectory policy across HuRoN \cite{Hirose2023SACSoNSA} and SCAND \cite{Karnan2022SociallyCN} to learn general navigation priors; Stage~2 fine-tunes FastNav on InternData-N1, conditioning trajectory generation on cached plan tokens from the frozen slow system. Models are trained on 8 $\times$ NVIDIA A800 GPUs using the AdamW \cite{Loshchilov2017DecoupledWD} optimizer, and online inference is evaluated on an NVIDIA RTX PRO 6000 GPU. 

\noindent\textbf{Benchmark \& Metrics.}
We evaluate HAVANA on OmniNavBench \cite{Sun2026BeyondIA} across wheeled, quadrupedal, and humanoid embodiments in physics-enabled Matterport3D \cite{Chang2017Matterport3DLF} and GRScene \cite{Wang2024GRUtopiaDG} environments. Following the official protocol, all methods are evaluated zero-shot using publicly released checkpoints, with only embodiment-specific camera settings and robot footprints adapted to each method. We focus on goal-directed navigation, reporting \textbf{OSR} (Oracle Success Rate), and \textbf{SPL} (Success weighted by Path Length) for goal attainment,  \textbf{CSR} (Completion Success Rate) and \textbf{SGC} (Sub-Goal Completion) indicate partial progress, while leaving human-aware interaction objectives outside the scope of this work.  Since HAVANA generates continuous trajectories without an explicit termination action, we use \textbf{OSR} as the primary success metric to isolate goal-reaching performance from stop prediction, which OmniNavBench identifies as a distinct failure mode. All metrics are averaged over the corresponding test episodes, using $N_{\text{task}}$ where applicable.

\noindent\textbf{Baselines.} We compare HAVANA with state-of-the-art methods that are both closely related to our task and reproducible with publicly available models. \textbf{DualVLN} provides the closest architectural comparison as a dual-system VLN framework, while \textbf{AstraNav-World} represents a recent world-model-based navigation approach with explicit future prediction. As AstraNav-World only releases its Action-Former variant, we adopt this version for comparison. All methods are evaluated zero-shot on the same OmniNavBench split without in-domain fine-tuning.
\subsection{Comparative Results}
\noindent\textbf{Cross-Embodiment Navigation.} Table~\ref{tab:main_results} compares the zero-shot transfer performance of HAVANA with DualVLN and AstraNav-World across three robot embodiments. HAVANA achieves the highest OSR, CSR, SGC, and SPL on all evaluated subsets, showing consistent gains across multiple measures of navigation performance. On the complete 411-episode Carter split, HAVANA reaches an OSR of $50.36\%$, exceeding DualVLN and AstraNav-World by $25.54$ and $36.98$ percentage points, respectively. On Aliengo, HAVANA achieves $27.43\%$ OSR, improving over the two baselines by $4.09$ and $8.27$ points. On H1, its OSR reaches $27.02\%$, compared with $6.96\%$ for DualVLN and $11.73\%$ for AstraNav-World. The improvements extend beyond final-goal reaching: on Carter, HAVANA increases CSR and SGC by $5.81$ and $7.35$ points over DualVLN, while achieving the highest SPL on all three embodiments. These concurrent gains indicate that improved goal reaching is accompanied by stronger subgoal completion and success-weighted path efficiency.

The gains nevertheless vary across embodiments. Improvements are more pronounced on Carter and H1, whereas Aliengo shows smaller margins, particularly in CSR and SGC, likely due to embodiment-specific dynamics and viewpoint shifts that increase subgoal-tracking noise. Overall, HAVANA delivers the strongest navigation performance across all three embodiments.

\begin{figure}[t]
    \centering
    \includegraphics[width=\linewidth]{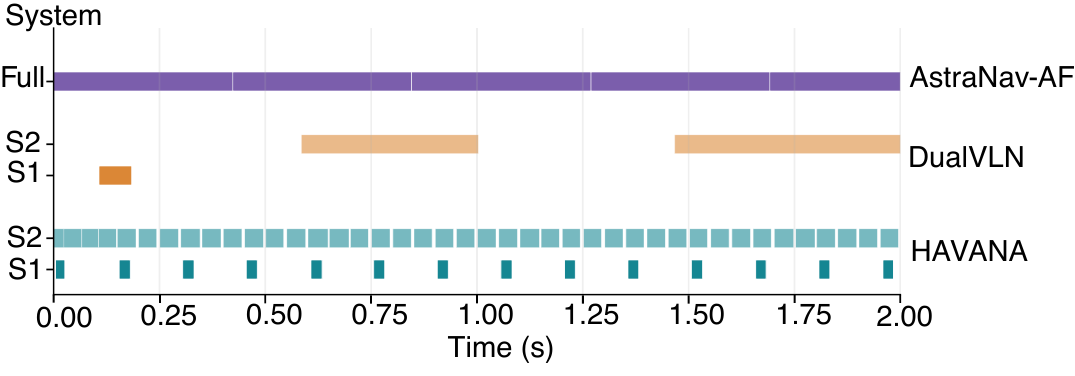}
    \caption{
         \textbf{Illustration of inference timelines.} Each rectangle marks the wall-clock interval of one inference call. AstraNav-AF is shown as a full decision pipeline, while HAVANA and DualVLN retain their native scheduling. S1 and S2 denote the slow and fast systems in each method.
    }
    \vspace{-1em}
    \label{fig:inference_timeline}
\end{figure}

\noindent\textbf{Inference Efficiency.} As shown in Figure \ref{fig:inference_timeline}, We measure inference latency after warm-up using one inference worker and an observation batch size of one. Tests use six fixed Carter cases, with each method retaining its original inputs and inference procedure. S1 and S2 refer to the slow planner and fast trajectory policy, respectively. HAVANA's slow planner has a median latency of 150.52 ms, compared with 424.97 ms for DualVLN's slow planner and 433.42 ms for AstraNav-AF's full decision pipeline. We compare the fast policies on the five shared cases where DualVLN produces the required conditioning latents. HAVANA's median and 95th-percentile latencies are 17.00 and 17.87 ms, respectively, compared with 73.42 and 80.43 ms for DualVLN. This speedup follows FastNav's design: it conditions on a compact set of cached plan tokens and predicts a short waypoint sequence with a lightweight 1D U-Net, avoiding long-horizon autoregressive decoding. Moreover, asynchronous scheduling overlaps S1 with S2 so control steps use the latest completed plan without blocking.

\subsection{Analysis of Predictive World Modeling}
\label{sec:4.3}

\begin{figure*}[t]
    \centering
    \includegraphics[width=\linewidth]{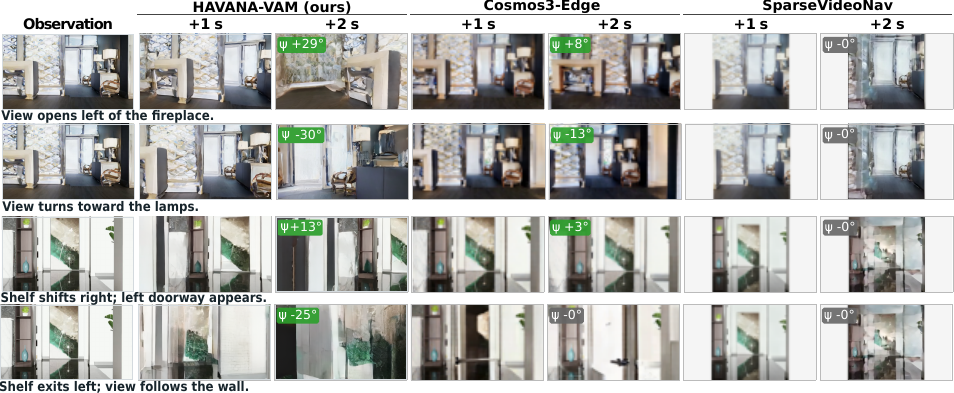}
    \caption{
         \textbf{Qualitative comparison of predictive world modeling.} The top two rows show Matterport3D scenes and the bottom two show GRScenes. Predictions are shown with signed heading changes at +1 s and +2 s. HAVANA exhibits more pronounced viewpoint changes under the turn instructions, while Cosmos3-Edge and SparseVideoNav better preserves visual fidelity and scene structure. We provide a diagnostic analysis of directional consistency in Sec. \ref{sec:4.3} .
    }
    \label{fig:world_modeling}
\end{figure*}

\begin{figure*}[t]
    \centering
    \includegraphics[width=\linewidth]{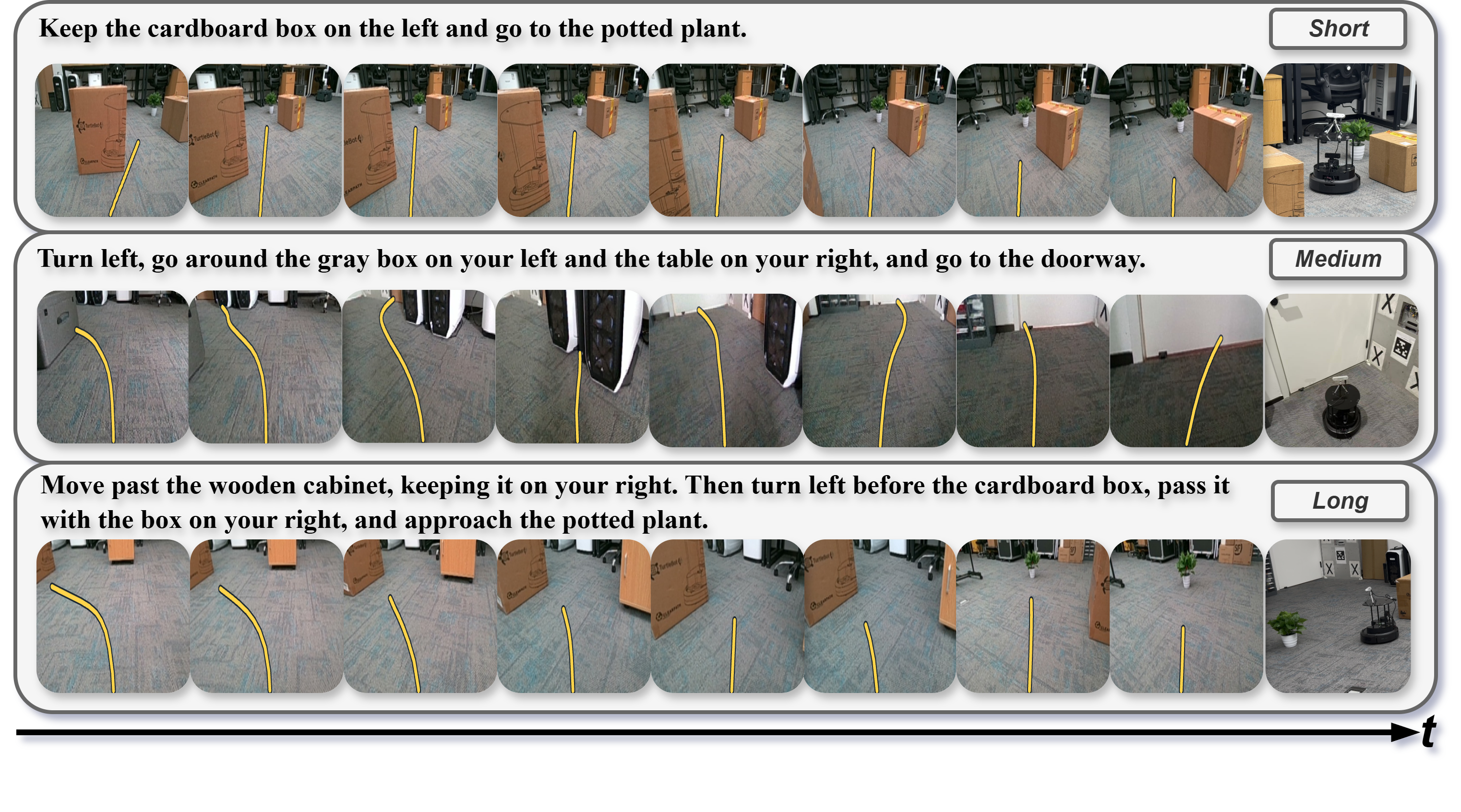}
    \vspace{-2em}
    \caption{
         \textbf{Ego-centric frames for zero-shot deployment in real-world scenarios with instructions of varying length.} At each step it predicts a short-horizon robot-centric trajectory conditioned on the current observation and latest latent plan. The trajectory is continuously planned as new observations arrive, with a controller selecting \textcolor{Dandelion}{executable trajectory} for closed-loop navigation. 
    }
    \vspace{-1em}
    \label{fig:real_robot_demo}
\end{figure*}
\noindent\textbf{Future Visual Prediction.}
We randomly sample eight Carter observation sequences from OmniNavBench and evaluate each model with paired left- and right-turn instructions through its supported inference interface, shown in Figure \ref{fig:world_modeling}. Cosmos3-Edge and SparseVideoNav produces videos with higher visual fidelity and better-preserved scene structure, whereas HAVANA exhibits more pronounced viewpoint changes in response to the turn instructions. To quantify motion in the predicted videos, we run a shared LingBot-Map estimator \cite{chen2026geometric} at a 2\,s prediction horizon. HAVANA yields a mean absolute heading change of $30.6^{\circ}$ and a depth-normalized camera displacement of 1.077, versus $6.7^{\circ}$ and 0.106 for Cosmos3-Edge and $0.089^{\circ}$ and 0.003 for SparseVideoNav. These observations highlight HAVANA’s stronger viewpoint responses under turn instructions. While Cosmos3-Edge and SparseVideoNav better preserves visual fidelity, HAVANA uses future-video prediction to supervise latent plans for trajectory generation, without decoding future videos during navigation.
\subsection{Real-world Deployment}

\noindent\textbf{Real-robot Setup.} We further evaluate HAVANA through zero-shot deployment on a wheeled robot across three instruction-length levels. Experiments are conducted in unseen indoor environments, with natural-language instructions involving directional turns, obstacle avoidance, and goal-directed navigation. At each control step, HAVANA predicts a short-horizon robot-centric trajectory from the current onboard observation and latest latent plan, followed by an MPPI controller that derives velocity commands for execution. The same navigation model is deployed without real-world fine-tuning, requiring only deployment-specific configuration adjustments.
\begin{figure*}[t]
    \centering
    \includegraphics[width=\linewidth]{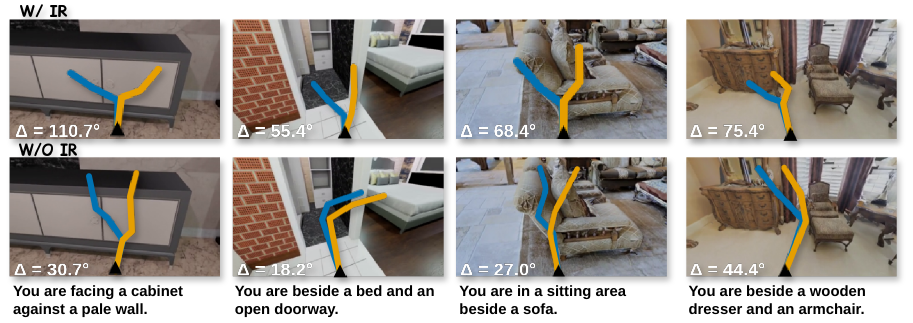}
    \vspace{-1em}
    \caption{
         \textbf{Effect of instruction-ranking (IR) on directional grounding.}
Each column shows the same observation queried with paired counterfactual instructions that differ only in the turn direction. The colored curves denote the corresponding predicted trajectories, and $\Delta$ measures the absolute difference between their endpoint yaw angles. Top: w/ IR; bottom: w/o IR. \textcolor[HTML]{2D66A5}{Blue: turn left};
\textcolor[HTML]{D69828}{orange: turn right}.
    }
    \label{fig:ir_ablation}
    \vspace{-1em}
\end{figure*}

\noindent\textbf{Analysis of Real-World Rollouts.} Figure \ref{fig:real_robot_demo} illustrates how HAVANA executes navigation instructions of increasing complexity through closed-loop trajectory updates. In the short-instruction example, the robot navigates past a cardboard box toward the target plant. The medium-instruction example requires a left turn while avoiding obstacles on both sides to reach the doorway. The long-instruction example involves multiple sequential constraints, including passing a cabinet, turning before a cardboard box, and approaching the target plant. Notably, the spatial extent of predicted trajectories adapts to visible free space, extending over open ground and shortening under constrained visibility or nearby obstacles. Across these rollouts, the local trajectories are repeatedly updated as the robot moves and new observations become available, rather than remaining fixed to an initial prediction. These results demonstrate how HAVANA translates instruction-conditioned reasoning into reactive local motion during real-world navigation.

\subsection{Ablation Studies}

\noindent\textbf{Denoising Objective.}
\input{tables/denoise_ablation}
We investigate the effect of future-video denoising supervision on waypoint and local trajectory prediction, as shown in Table \ref{tab:denoise_ablation}. Denoising improves long-horizon waypoint prediction (0.76\%), suggesting benefits from modeling beyond-the-view changes. More importantly, denoising yields only marginal gains with fresh plans (0.81\%) but substantially reduces trajectory ADE under cached plans (8.84\%), narrowing the performance gap between fresh and delayed predictions. This finding is consistent with our motivation for learning short-horizon reactive motion from long-horizon predictive latents: visual world-modeling supervision helps preserve useful motion guidance when planning updates are delayed. Together with the strong closed-loop navigation results in Table \ref{tab:main_results}, these observations suggest that the value of predictive world modeling using a denoising objective.

\noindent\textbf{Instruction-Ranking Objective.} As demonstrated in Table \ref{tab:ir_ablation}, we evaluate the instruction-ranking objective using paired counterfactual prompts that keep the scene description fixed while changing only the turn direction. On 64 paired observations, the directional response accuracy increases from 71.09\% to 86.72\%, while the proportion of cases in which both left and right variants are predicted correctly under the same observation rises from 42.19\% to 73.44\%. As shown in Figure \ref{fig:ir_ablation}, without instruction-ranking supervision the predicted futures often remain visually similar under opposite commands, indicating weak directional sensitivity. Adding the instruction-ranking objective yields clearly separated future predictions and more consistent directional responses. 
\input{tables/ir_ablation}

%% file: tables/omninavbench.tex
\begin{table*}[t]
    \centering
    \caption{\textbf{Quantitative comparison with baselines across different robot embodiments.}
    All metrics are reported in \%.}
    \label{tab:main_results}

    \footnotesize
    \renewcommand{\arraystretch}{1.15}
    \setlength{\tabcolsep}{3pt}

    \begin{tabularx}{\textwidth}{
        @{}
        >{\raggedright\arraybackslash}p{31mm}
        >{\raggedright\arraybackslash}p{27mm}
        *{4}{>{\centering\arraybackslash}X}
        @{}
    }
    \toprule

    \textbf{Method} &
    \textbf{Embodiment} &
    \textbf{OSR}$\uparrow$ &
    \textbf{CSR}$\uparrow$ &
    \textbf{SGC}$\uparrow$ &
    \textbf{SPL}$\uparrow$
    \\
    \midrule

    \multirow{3}{*}{
        \shortstack[l]{
        \textbf{AstraNav-World}\\
        (w/ Action Former)
        }
    }
    & NVIDIA Carter
    & 13.38 & 19.70 & 36.94 & 9.94\\

    & Unitree Aliengo
    & 19.16 & 22.79 & 39.86 & 11.57 \\

    & Unitree H1
    & 11.73 & 18.59 & 36.60 & 9.46 \\

    \midrule

    \multirow{3}{*}{\textbf{DualVLN}}
    & NVIDIA Carter
    & 24.82 & 23.34 & 40.84 & 18.28 \\

    & Unitree Aliengo
    & 23.34 & 23.82 & 41.28 & 16.21 \\

    & Unitree H1
    & 6.96 & 13.03 & 30.74 & 5.08 \\

    \midrule
    
    \multirow{3}{*}{\textbf{Ours}}
    & NVIDIA Carter
    & \textbf{50.36} & \textbf{29.15}
    & \textbf{48.19} & \textbf{26.43}  \\

    & Unitree Aliengo
    & \textbf{27.43} & \textbf{24.54}
    & \textbf{41.51} & \textbf{17.83} \\

    & Unitree H1
    & \textbf{27.02} & \textbf{26.02}
    & \textbf{45.78} & \textbf{25.03} \\

    \bottomrule
    \end{tabularx}
\end{table*}

%% file: tables/denoise_ablation.tex
\begin{table}[t]
    \centering
    \caption{\textbf{Effect of future-video denoising supervision on offline prediction.} We report Average Displacement Error (ADE~$\downarrow$) for long-horizon waypoints and short-horizon trajectories. \emph{Cached} evaluates previously generated latent plans, and $\Delta$ denotes relative improvement over the baseline without denoising.}
    \label{tab:denoise_ablation}
    \footnotesize
    \renewcommand{\arraystretch}{1.15}

    \begin{tabular*}{\columnwidth}{
        @{\extracolsep{\fill}}
        lccc
        @{}
    }
    \toprule
    \textbf{Metric} &
    \textbf{w/o Denoise} &
    \textbf{Denoise} &
    \textbf{$\Delta$} \\
    \midrule
    Waypoint ADE      & 0.3891          & \textbf{0.3861} & +0.76\% \\
    Fast Traj. ADE (Fresh)  & 0.1671          & \textbf{0.1658} & +0.81\% \\
    Fast Traj. ADE (Cached) & 0.2321          & \textbf{0.2116} & +8.84\% \\

    \bottomrule
    \end{tabular*}
\end{table}

%% file: tables/ir_ablation.tex
\begin{table}[t]
    \centering
    \caption{\textbf{Effect of instruction-ranking (IR) supervision on directional grounding.} \emph{Directional accuracy}~($\uparrow$) measures whether the endpoint yaw follows the instructed direction, while \emph{Both variants correct}~($\uparrow$) requires correct predictions for both counterfactual prompts on the same observation.}
    \label{tab:ir_ablation}

    \footnotesize
    \renewcommand{\arraystretch}{1.15}

    \begin{tabular*}{\columnwidth}{
        @{\extracolsep{\fill}}
        lcc
        @{}
    }
    \toprule
    \textbf{Metric} &
    \textbf{w/o IR} &
    \textbf{w/ IR} \\
    \midrule

    Directional accuracy   & 0.7109 & \textbf{0.8672} \\
    Both variants correct  & 0.4219 & \textbf{0.7344} \\

    \bottomrule
    \end{tabular*}

\end{table}

%% file: sec/5_conclusion.tex
\section{Conclusion}
In this work, we presented HAVANA, a dual-system VAM for VLN that decouples predictive world modeling from reactive trajectory prediction. By transferring instruction-conditioned compact future representations to a reactive policy, HAVANA enables asynchronous closed-loop navigation. Experiments on OmniNavBench demonstrate superior goal-reaching performance over competitive baselines across three robot embodiments. Zero-shot deployments on real robots further demonstrate its real-world transferability. Future work will focus on the generalization across robotic platforms in unstructured real-world environments.

\section*{Acknowledgements}
The computations in this paper were supported by Tongji University Intelligent Computing Center.